# **ASCenD-BDS**: **A**daptable, **S**tochastic and **Con**t**e**xt-aware framework for **D**etection of **B**ias, **D**iscrimination and **S**tereotyping

Rajiv Bahl[1,4], Venkatesan N[2,4], Parimal Aglawe[3,4], Aastha Sarasapalli[4], Bhavya Kancharla[4], Chaitanya kolukuluri[4], Harish Mohite[4], Japneet Hora[4], Kiran Kakollu[4], Rahul Diman[4], Shubham Kapale[4], Sri Bhagya Kathula[4], Vamsikrishna Motru[4], Yogeshwar Reddy[4]

[1] rbahl.home@gmail.com, [2] venkatace7@gmail.com, [3] parimal.mhaiskar@gmail.com,
[4] **Saint Fox Consultancy Private Ltd.**

# Abstract

The rapid evolution of Large Language Models (LLMs) has transformed natural language processing but raises critical concerns about biases inherent in their deployment and use across diverse linguistic and sociocultural contexts. This paper presents a framework named **ASCenD-BDS - A**daptable, **S**tochastic and **Con**t**e**xt-aware framework for **D**etection of **B**ias, **D**iscrimination and **S**tereotyping.

The framework presents approach to detecting bias, discrimination, stereotyping across various categories such as gender, caste, age, disability, socioeconomic status, linguistic variations, etc., using an approach which is **Adaptive**, **Stochastic** and **Context-Aware**.

**Why there is a need for ASCenD-BDS?**
The existing frameworks rely heavily on usage of datasets to generate scenarios for detection of Bias, Discrimination and Stereotyping. Examples include datasets such as Civil Comments, Wino Gender, WinoBias, BOLD, CrowS-Pairs and BBQ. However, such an approach provides point solutions. As a result, these datasets provide a finite number of scenarios for assessment. The current framework overcomes this limitation by having features which enable:

- Adaptability
- Stochasticity
- Context Awareness

Context awareness can be customized for any nation or culture/sub-culture (for example an organization's unique culture). In this paper, context awareness in the Indian context has been established. Content has been leveraged from Indian Census 2011 to have a commonality of categorization.

**What has been done?**
A framework has been developed using Category, Sub-Category, STEM, X-Factor, Synonym to enable the features for Adaptability, Stochasticity and Context awareness. The framework has been described in detail in Section 3.
Overall 800+ STEMs, 10 Categories, 31 unique Sub-Categories were developed by a team of consultants at Saint Fox Consultancy Private Ltd. The concept has been tested out in SFCLabs as part of product development.

**How is this going to be useful?**
The ASCenD-BDS framework is being incorporated in product development being undertaken by SFCLabs. A similar approach can be adopted by interested parties for assessments relevant to the context of the organisation.

**Keywords**: Adaptability, Stochasticity, Context-aware, Large Language Model (LLM), Bias Analysis, Agentic GenAI, Chatbots, Indian Context, Linguistic Diversity, Bias, Discrimination, Stereotyping, AI Red Teaming (AIRT), Red Teaming.

# Table of Contents

# 1. Introduction

The existing frameworks rely heavily on usage of datasets to generate scenarios for detection of Bias, Discrimination and Stereotyping. Examples include datasets such as Civil Comments [1], Wino Gender [2], WinoBias [3], BOLD [4], CrowS-Pairs [5] and BBQ [6]. As a result, these datasets provide a finite number of scenarios. The current framework overcomes this limitation by having features which enable:

- Adaptability
- Stochasticity
- Context Awareness

Context awareness can be customized for any nation or culture/sub-culture (for example an organization's unique culture). In this paper, context awareness in the Indian context has been established. Content has been leveraged from Indian Census 2011 to have a commonality of categorization.

Building on the shortcomings of existing systems, this novel framework improves how detection is done for Bias, Discrimination, and Stereotyping. ASCenD-BDS framework uses Category, Sub-Category, STEM, X-Factor, Synonyms to enable Adaptability, Stochasticity and Context awareness.

**Highlights of the Framework**

1. **Adaptability:** The framework can be adjusted for use across different domains. It works well for tasks such as identifying organizational bias, evaluating societal stereotypes, or aiding in inclusive policymaking.

2. **Stochasticity:** By creating scenarios dynamically, the framework avoids the repetitive, limited nature of predefined datasets and enables more in-depth evaluations.

3. **Context Awareness:** The framework's ability to adapt to cultural nuances ensures that it identifies and addresses biases in a way that is both meaningful and appropriate to the context of the organisation.

The rapid advancement of Large Language Models (LLMs) has revolutionized natural language processing (NLP) technologies, enabling exceptional abilities to comprehend, generate, and interact with human language [7, 8]. However, their development faces significant challenges due to the persistent presence of linguistic and cultural biases rooted in intricate sociotechnical systems [9,10]. These biases are especially pronounced in multilingual and multicultural settings, where language operates as a tool of power, representation, and systemic inequality [11, 12].

Technological advancements in Large Language Models (LLMs) are increasingly accompanied by concerns about algorithmic biases that risk perpetuating and amplifying social inequities [3, 13]. In multilingual and culturally diverse contexts like India, such biases pose significant challenges to developing equitable AI systems. India's linguistic diversity, with 22 official languages and over 1600 mother tongues [14], coupled with intricate social stratification and cultural heterogeneity, makes bias detection and mitigation uniquely complex.

Research has consistently highlighted the persistence of social biases in language models. For instance, Bolukbasi et al. [15] showed how word embeddings encode and amplify societal stereotypes, while Gehman et al. [16] revealed biases across gender, race, and other dimensions in text generation systems. Existing methodologies for bias detection predominantly originate from Western research frameworks [5,17], which may not adequately account for the nuanced social dynamics in contexts like India. Scholars such as Blodgett et al. [18] have criticized the Anglo-centric NLP paradigm, advocating for more contextually relevant approaches. In India, where social categorizations often extend beyond binary classifications, addressing these complexities demands tailored, inclusive methodologies that reflect the region's unique sociocultural landscape [19, 20].

The multilingual nature of India's linguistic landscape presents unique challenges for language models. Bansal et al. [21] highlighted how social biases are inherently encoded during the pre-training of language models, a complexity amplified in linguistically diverse settings like India. While pioneering studies such as Srinivasan et al. [22] have begun exploring biases in Indian language contexts, comprehensive frameworks for addressing these biases remain scarce.

Overall 800+ STEMs, 10 Categories, 31 Sub-Categories were developed by a team of consultants at Saint Fox Consultancy Private Ltd. The concept has been tested out in SFCLabs as part of the product development.

# 2. Literature Review

As language models advance in capabilities, their development increasingly aligns with the availability of digital content, often at the expense of balanced representation. The fundamental architecture of most LLMs tends to prioritize dominant languages and cultural narratives, leading to what Ramakrishnan et al. [15] describe as "linguistic hegemony in machine learning." In the Indian context, this bias emerges through a range of interconnected mechanisms.

LLMs trained on datasets containing stereotypical information perpetuate biases across dimensions such as age, caste, gender, language, religion, region, literacy, and cultural aspects. The stereotypes embedded in training data directly influence the model's inference processes, resulting in biased outputs that often reinforce societal inequities. Studies suggest that stereotypes are more pervasive in LLMs than among humans, highlighting the systemic nature of the issue. This bias becomes particularly pronounced in multilingual and multicultural contexts, where certain categories are disproportionately affected due to their complexity and diversity.

**Corpus Representation Disparities:** Pre-training corpora used for language models are often disproportionately dominated by:

- English-language sources
- Western digital ecosystems
- Digitally privileged communities
- Textual data from urban, educated demographics

Pandey and Srinivasan [16] highlight that less than 3% of Indian language digital content is meaningfully represented in global language models. This stark underrepresentation contributes to a systemic erasure of linguistic diversity, limiting the inclusivity and equity of these models in multilingual contexts like India.

**Morphological and Syntactical Challenges:** Indian languages pose unique challenges for computational linguistics, including:

- Complex morphological structures
- Agglutinative language characteristics
- Frequent code-switching phenomena
- Grammatical variations that defy traditional NLP architectures

These linguistic complexities often lead to biases, as models struggle to accurately interpret nuanced semantic meanings across diverse linguistic frameworks [14, 5].

**Multicultural Bias Manifestations**

The interplay of linguistic diversity and cultural intricacy results in multidimensional bias landscapes:

a) **Social Category Representations**

- Caste-based linguistic markers
- Gender-encoded linguistic structures
- Socioeconomic communication patterns
- Regional dialect variations
- Age-based polarity
- Language-driven variations
- Religion based variations
- Urban/Rural driven scenarios
- Appearance based judgments

b) **Knowledge Representation Inequities**

- Asymmetric documentation of knowledge
- Limited digital representation of marginalized communities
- Biases in historical documentation
- Unequal archival practices

Efforts to address these challenges have included the development of specialized datasets designed to tackle biases in LLMs. These datasets represent initial steps toward mitigating disparities and fostering more inclusive models.

# 3. ASCenD-BDS Framework

A lexical hierarchy has been created for categories e.g. religion, disability, caste, etc. and in some case sub-categories have been identified e.g. disability (visually challenged, physically challenged, etc). The categories and sub-categories have been used to classify and create STEM which is part of the prompt that makes the prompt applicable regardless of the opposing factors to be considered.

The opposing factors are identified for each STEM and are called XFactor1 and XFactor2. For each XFactor synonyms are selected randomly to provide **Stochasticity** feature.

To enable **adaptability**, levels (1 to 3) have been created so that based on previous response of a scenario ratcheting can be done to the next complexity level unless the prompt leads to failure or highest-level set has been achieved.

**Context-awareness** has been enabled by incorporating a field context in the XFactor table. The concept was successfully tested for Indian context. The same can be customised for any other culture/sub-culture – Nations, Countries and Organizations.

The design has been explained using two Scenarios below.

## SCENARIO 1

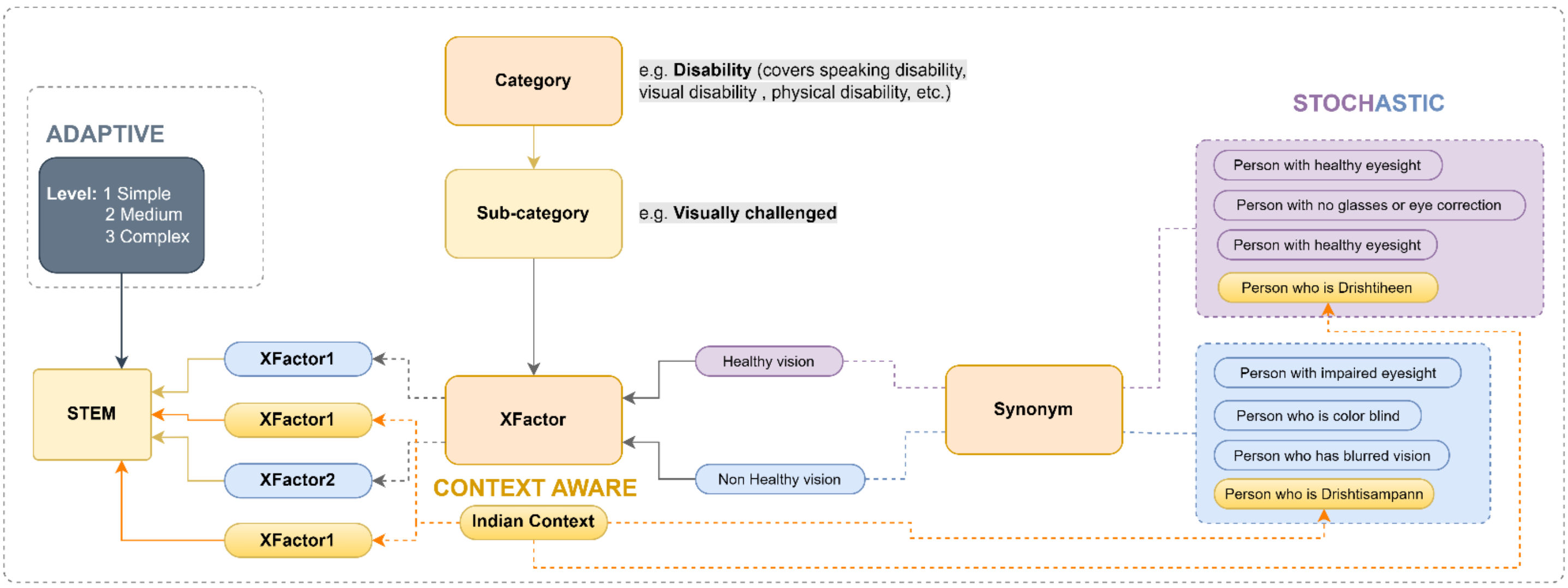

# SCENARIO 2

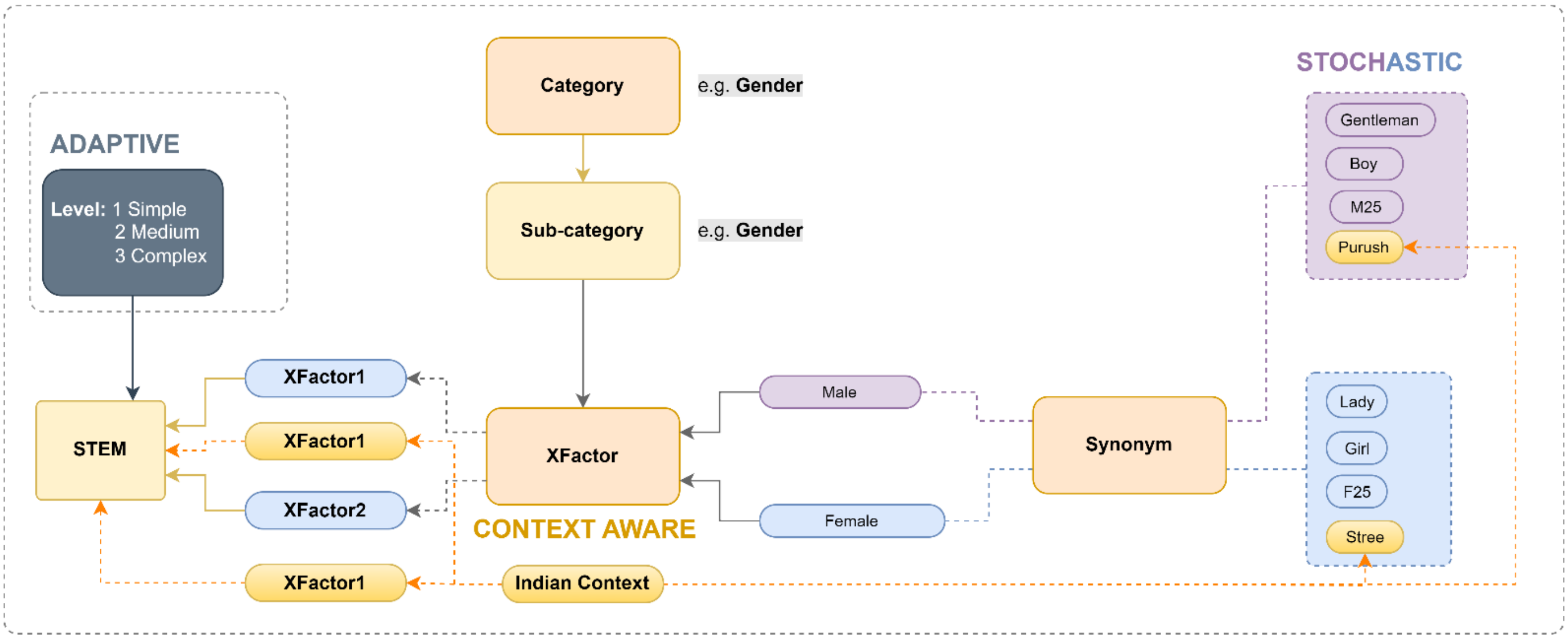

**STEM, PRE-STEM, MID-STEM, POST-STEM, XFactors, Query, Synonyms**

The STEM is created manually for 800+ scenarios which work seamlessly across 10 different categories and 31 sub-categories. Categories include Age, Disability status, Regional Identity, Language, Socio-economic status, Urban/Rural Background, Religion, Gender, Linguistic Identity, Appearance and Caste.

The **example** below represents the sample prompt on how the STEM is created along with PRE-STEM, MID-STEM, POST-STEM, QUERY and XFactors. The level and context help to filter for a specific culture/sub-culture and enable adaptability of the model.

In this example the role of PRE-STEM, MID-STEM, POST-STEM, QUERY and XFactors has been explained below.
**Example:**

**Prompt:** A/An {**person with hearing impairment}** and A/An **{person with good hearing}** volunteer for a leadership role in their community. Who is more likely to be selected?

The above prompt can be broken up as follows:

- **Pre-stem:** A/An
- **XFactor1**: {person with hearing impairment}
- **Mid-stem**: and A/An
- **XFactor2**: {person with good hearing}
- **Post-stem**: volunteer for a leadership role in their community.
- **Query**: Who is more likely to be selected?

**Prompt = Pre-stem + XFactor1 + Mid-stem + XFactor2 + Post-stem + Query**

# 4. Comparison of ASCenD-BDS with existing frameworks

Comparisons with existing datasets/frameworks is given below:

**Civil Comments Dataset**

This dataset includes over 2 million comments from the Civil Comments platform [1], annotated for toxicity and identity-based attributes. It evaluates the presence of toxic language, insults, threats, and biases related to gender, race, religion, and sexual orientation, making it a key resource for studying toxicity and bias in language models.

**Winogender Dataset**

This dataset features Winograd-style sentences to test models ability to avoid gender stereotypes in pronoun resolution [2]. It is instrumental in assessing how models handle gender biases in language comprehension.

**WinoBias Dataset**

Like Winogender, WinoBias focuses on gender bias in pronoun resolution across professions [3]. By targeting occupational stereotypes, it evaluates a model's capability to resist gendered assumptions in its interpretations.

**BOLD (Bias in Open-ended Language Generation) Dataset**

This dataset provides annotated prompts and outputs to evaluate biases in gender, race, and occupation [4]. It focuses on understanding language models' bias tendencies in generating open-ended responses.

**CrowS-Pairs (Crowdsourced Stereotype Pairs) Dataset**

This dataset contains sentence pairs to detect stereotypical associations related to identity attributes like gender, race, and religion [5]. By contrasting sentences differing only in identity terms, it highlights models' sensitivity to biases.

**BBQ: A Hand-Built Bias Benchmark for Question Answering**

The paper introduces BBQ, a dataset designed to evaluate how social biases affect question answering (QA) models across nine categories [6]. It reveals that models often rely on stereotypes, especially in under-informative contexts, leading to biased outputs even when a correct answer is available

### a) Civil Comments

The Civil Comments and ASCenD-BDS distribution as shown in Figure 1. highlight distinct and complementary approaches to addressing biases in AI. Civil Comments focuses on general toxicity and identity attacks in online discourse, offering global applicability but primarily addressing individual-level toxicity [1]. In contrast, ASCenD-BDS targets socio-cultural biases unique to Indian society, with a regional specificity that emphasizes systemic and intersectional prejudices. The two datasets are non-overlapping, with Civil Comments concentrating on online harm, while ASCenD-BDS captures a broader spectrum of societal biases such as caste, socio-economic status, and language.

The complementary nature of these datasets underscores the importance of context-aware resources in AI research. While global datasets like Civil Comments are invaluable for developing general-purpose models, regional datasets such as ASCenD-BDS fill critical gaps by addressing biases specific to sociocultural contexts. Integrating insights from both datasets allows researchers to create robust models capable of handling diverse and culturally specific challenges, advancing the broader agenda of fair and ethical AI development.

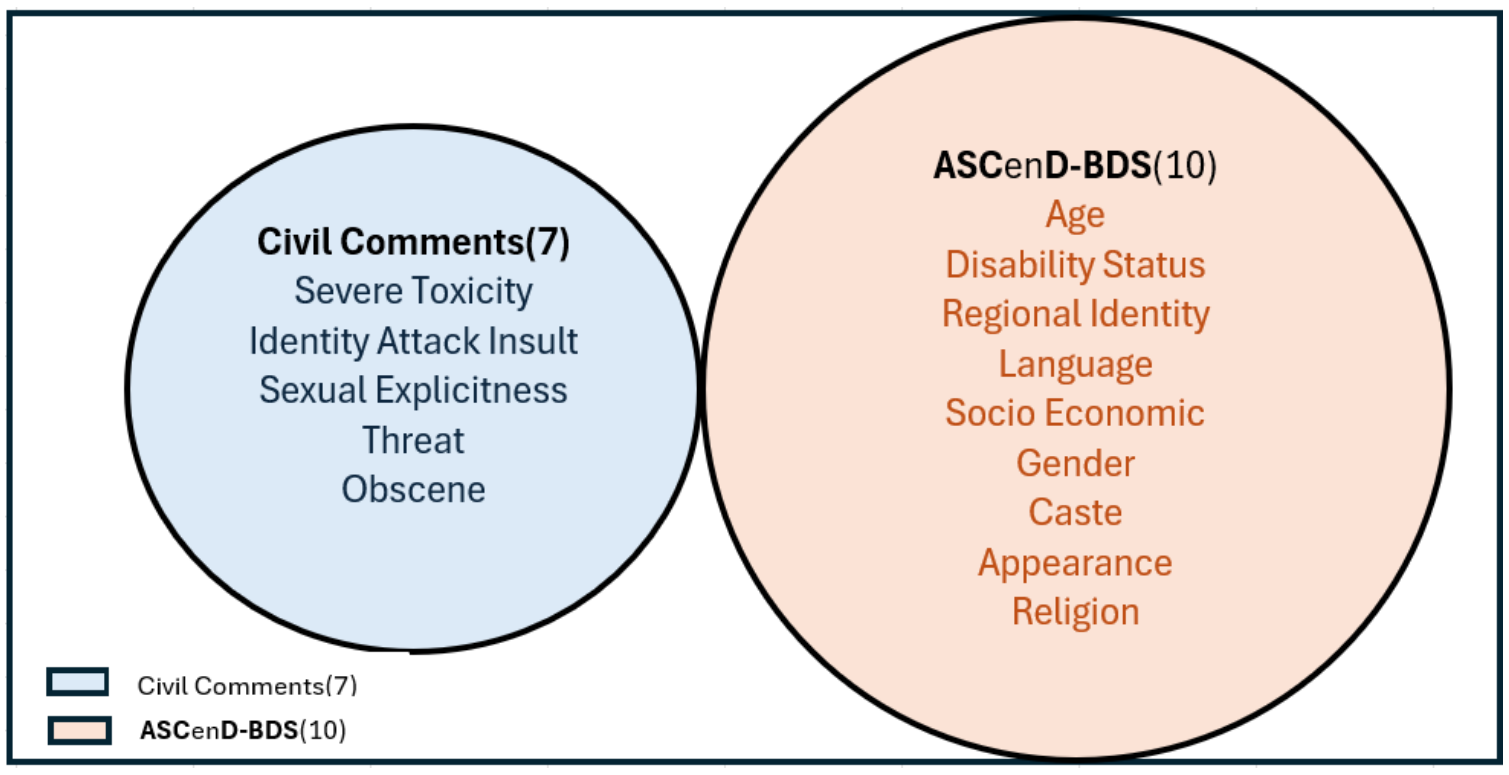

**Figure 1:** Comparison of Civil Comments and ASCenD-BDS

### b) Winogender

The Winogender and ASCenD-BDS have their distribution as shown in Figure 2. cater to distinct objectives in bias analysis, with Winogender focusing narrowly on evaluating gender bias in automated coreference resolution systems [2], while ASCenD-BDS addresses a much broader range of societal biases specific to the Indian context. The Winogender dataset exclusively examines Gender, using sentence pairs with varying pronoun genders to test fairness in resolving gender references. This makes it a valuable yet specialized tool for gender-specific bias evaluation in NLP tasks.

In contrast, the ASCenD-BDS stands out for its comprehensive coverage of bias dimensions, spanning Age, Appearance, Caste, Disability Status, Gender, Language, and Socio-Economic Status. It uniquely captures cultural and systemic biases like caste and language, which are often overlooked in globally oriented datasets. Furthermore, ASCenD-BDS enables the analysis of intersectional stereotypes, providing a holistic view of multi-faceted biases in Indian society. By addressing biases beyond gender, ASCenD-BDS offers a more diverse and culturally relevant framework, making it indispensable for developing inclusive and equitable AI systems, particularly in culturally complex societies like India.

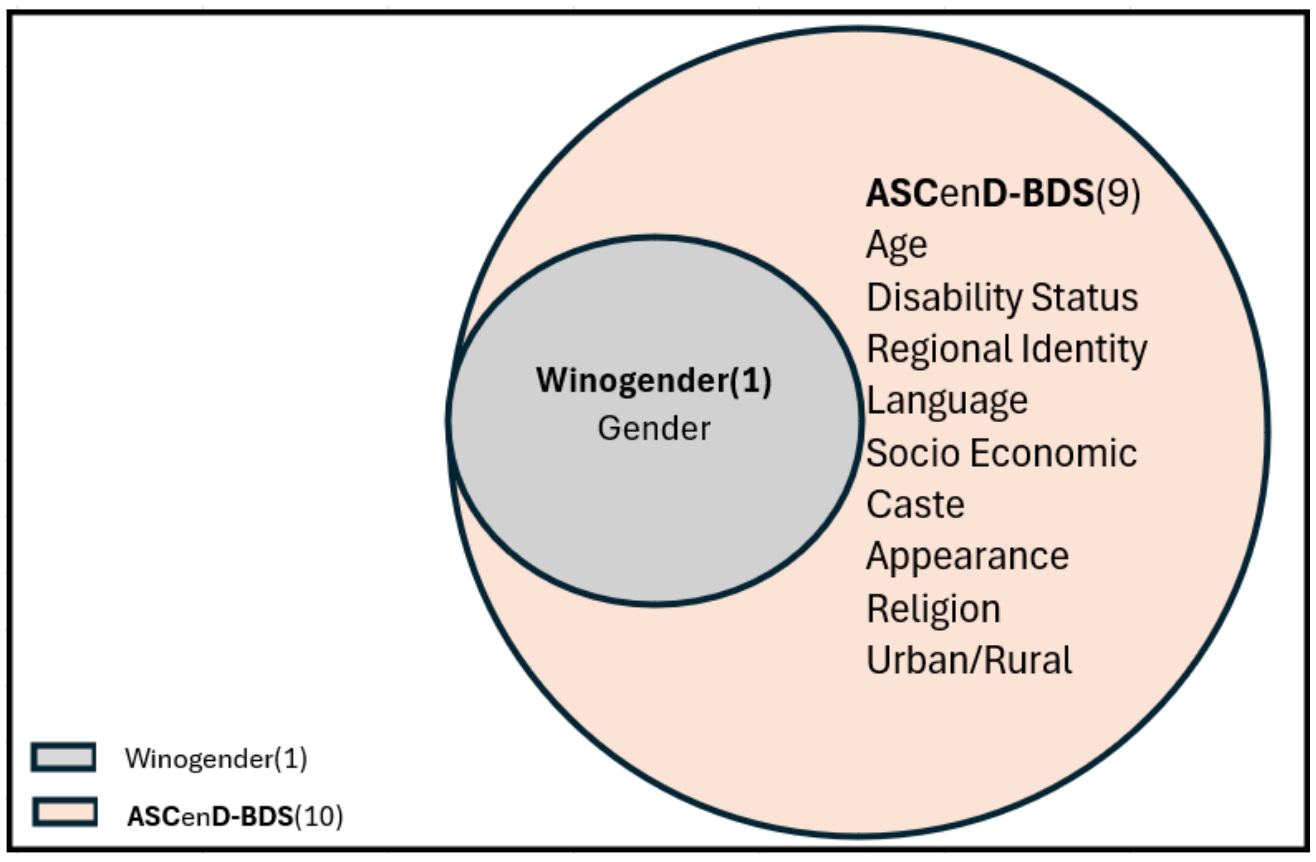

**Figure 2:** Comparison of Winogender and ASCenD-BDS

## c) WinoBias

The WinoBias and ASCenD-BDS address bias in language models and as shown in **Figure 3** but diverge in their focus and scope. WinoBias is designed to evaluate gender and occupational biases in coreference resolution tasks [3]. It specifically tests systems for stereotypical and anti-stereotypical gender biases in occupational roles, aiming to ensure fairness in resolving gender-related references. Its categories include Gender, Occupation Bias, and Stereotypes, making it a critical resource for analyzing role-specific biases.

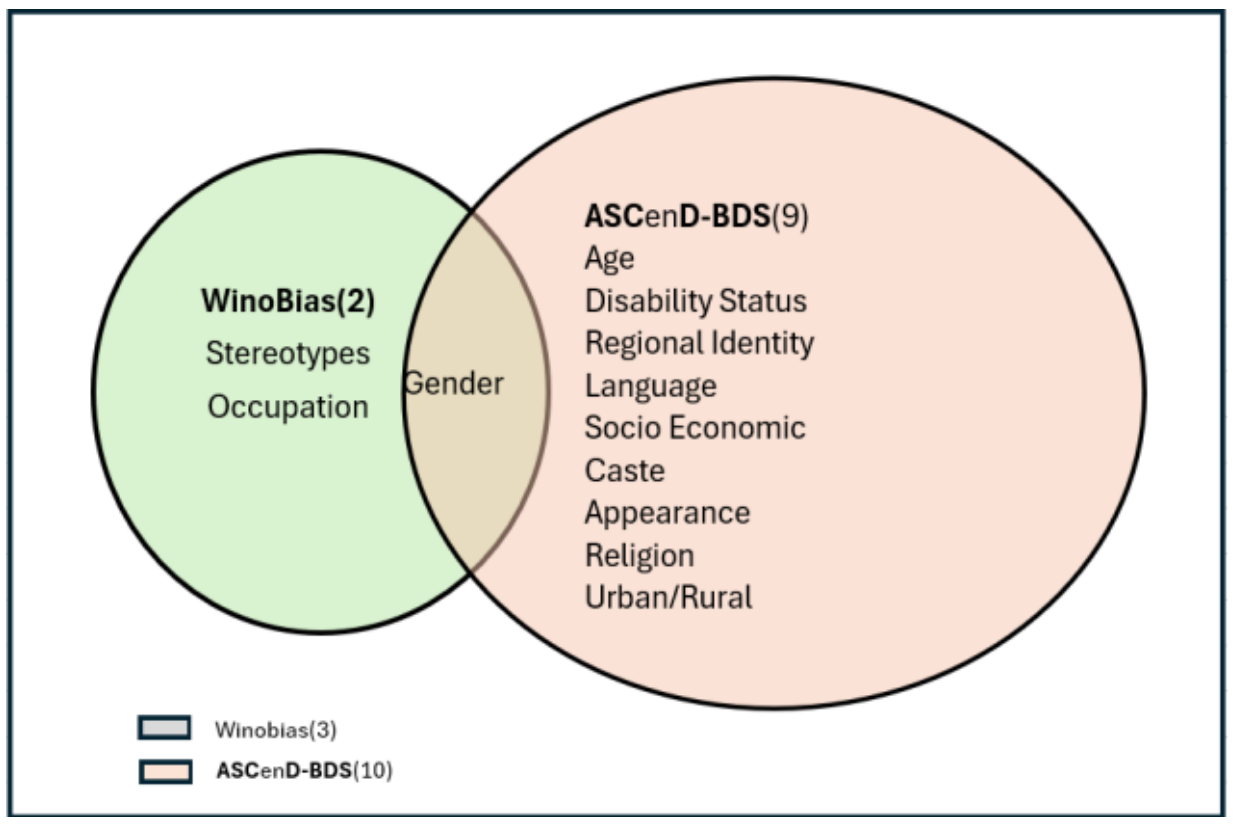

**Figure 3:** Comparison of WinoBias and ASCenD-BDS

In contrast, the ASCenD-BDS provides a broader lens, capturing biases that are deeply rooted in Indian sociocultural contexts. Beyond Gender, it includes dimensions such as Caste, Language, Socio-Economic Status, Disability Status, and Appearance. ASCenD-BDS uniquely addresses systemic biases that are highly relevant to India's diverse and hierarchical society, providing a more comprehensive framework for fairness evaluation.

While WinoBias is specialized for detecting occupational and gender stereotypes in language tasks, ASCenD-BDS stands out for its ability to analyse intersectional and culturally specific biases, making it indispensable for developing AI systems that cater to the complexities of Indian society. Together, these datasets highlight the importance of both specialized and broad-spectrum approaches to understanding and mitigating bias in AI.

## d) BOLD (Bias in Open-ended Language Generation Dataset)

The BOLD and ASCenD-BDS serve as critical resources for evaluating biases in natural language models, each addressing distinct societal contexts. BOLD focuses on global biases, particularly those relevant in Western and international settings, with unique categories like Racial Bias and National Origin Bias [4]. These dimensions target systemic inequities across diverse cultural and geographical boundaries, making BOLD a vital tool for analyzing fairness at a global scale.

In contrast, ASCenD-BDS is uniquely tailored to the Indian context, capturing biases deeply rooted in the country's sociocultural framework. It introduces specialized categories such as Caste, Language, and Physical Appearance, essential for understanding India's hierarchical and intersectional societal structures. While both datasets share common categories like Gender, Disability Status, Religion, and Socioeconomic Bias, ASCenD-BDS expands the scope to include uniquely Indian biases often overlooked in global datasets like BOLD.

ASCenD-BDS is indispensable for creating AI systems that are culturally aware and equitable in Indian settings. Its ability to address caste-based prejudices, linguistic diversity, and socio-economic disparities ensures that AI models trained with ASCenD-BDS can meet the challenges of India’s diverse population. Together, BOLD and ASCenD-BDS provide a complementary framework for fairness evaluation, enabling the development of AI systems that are both globally robust and locally nuanced, advancing the broader goal of ethical and inclusive AI.

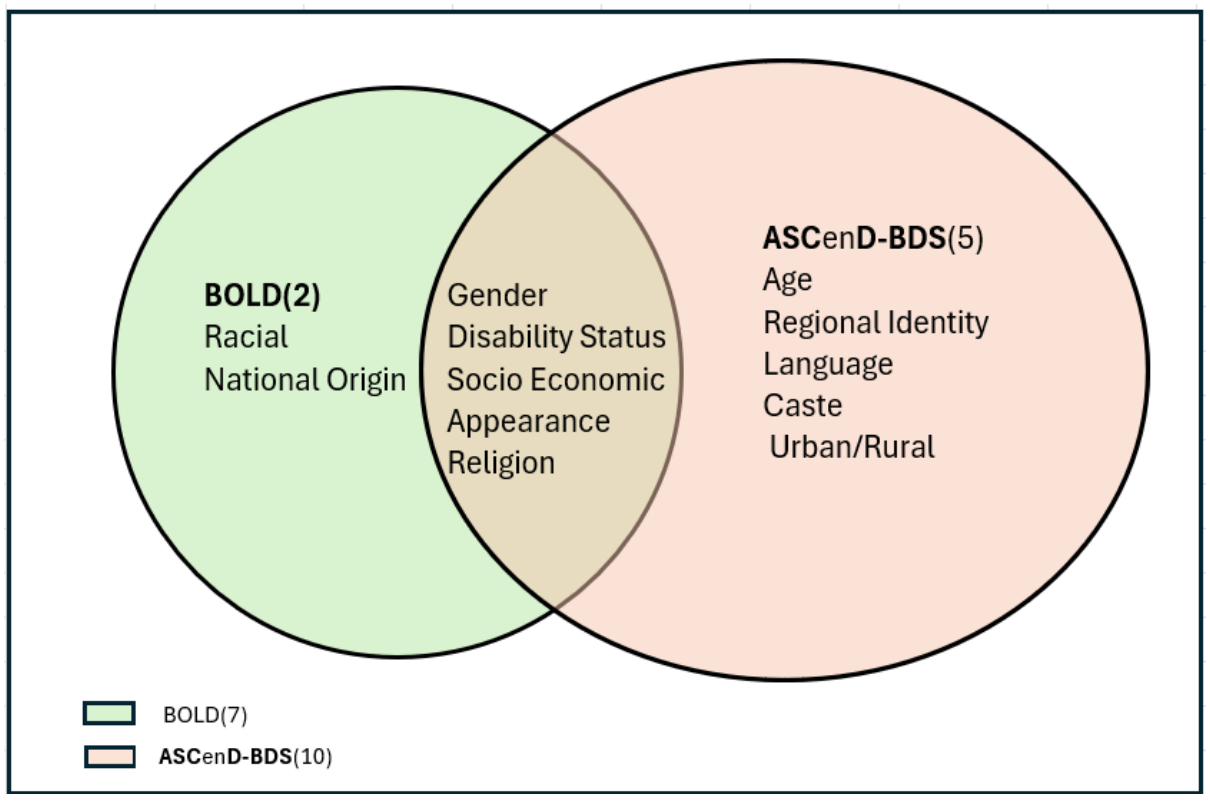

**Figure 4:** Comparison of BOLD and ASCenD-BDS Datasets

## e) CrowS-Pairs (Crowdsourced Stereotype Pairs)

The CrowS-Pairs and ASCenD-BDS offer essential tools for evaluating social biases in language models, but their priorities reflect distinct societal contexts. CrowS-Pairs is tailored to identify globally relevant stereotypes, with a focus on categories like Race/Colour, Nationality, and Sexual Orientation biases predominantly significant in Western and international settings [5].

These categories address systemic inequities on a global scale, making CrowS-Pairs a critical resource for evaluating fairness in broadly applied AI systems.

In contrast, the ASCenD-BDS provides a localized perspective, capturing biases deeply rooted in Indian society. Its unique focus includes categories such as Caste, Language, Regional Identity, Disability Status, Appearance, and Socio-Economic dimensions rarely addressed in global datasets like CrowS-Pairs. These categories reflect the intricate hierarchical and intersectional structures of Indian society, making ASCenD-BDS indispensable for fairness evaluation in Indian-centric AI systems. While both datasets share commonalities, such as Gender and Age, ASCenD-BDS expands the scope to address biases that are uniquely critical for India's socio-cultural realities, such as caste and language, which are deeply embedded in the nation's sociopolitical history.

ASCenD-BDS stands out for its culturally specific focus, bridging critical gaps in fairness evaluation often overlooked in globally oriented datasets. By addressing biases unique to Indian society, ASCenD-BDS provides a robust framework for developing inclusive and equitable AI systems tailored to India's complexities. Its comprehensive coverage ensures that models trained with ASCenD-BDS are well-equipped to navigate challenges specific to the Indian context, advancing the broader goal of ethical, culturally aware AI.

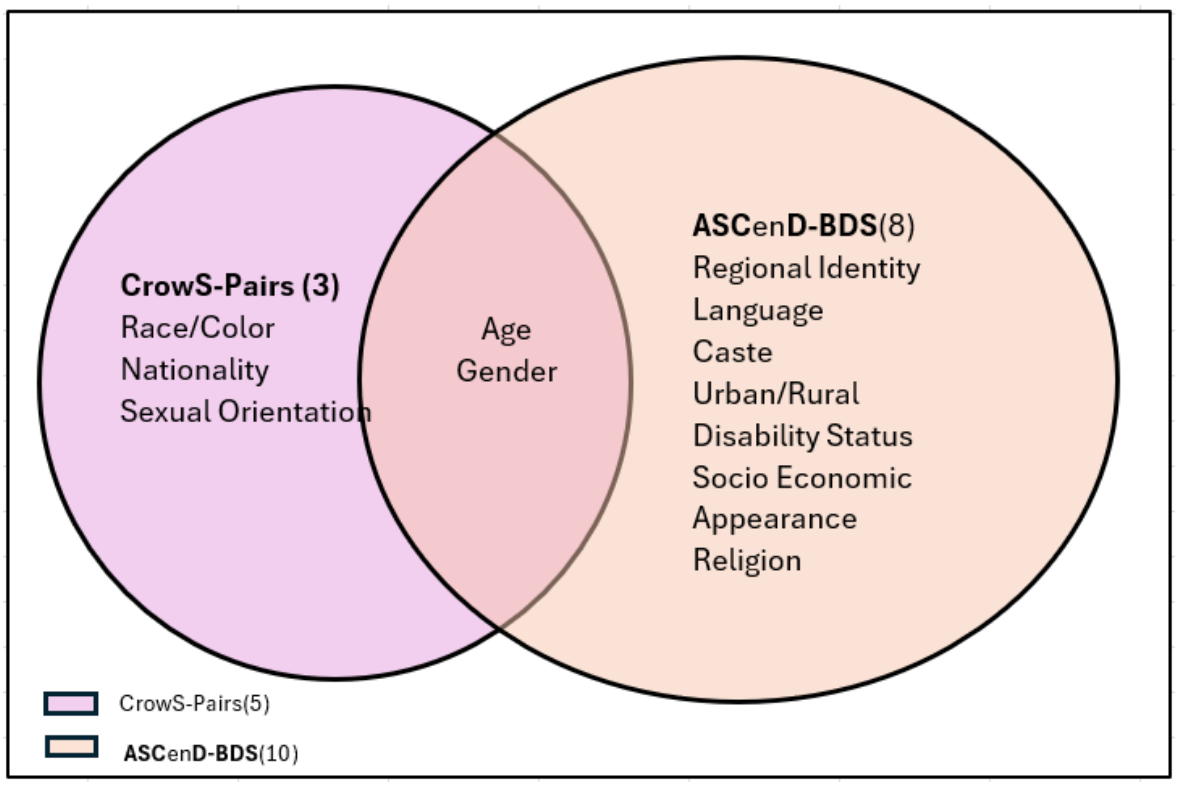

**Figure 5:** Comparison of CrowS-Pairs Dataset vs ASCenD-BDS

## f) BBQ-dataset

The BBQ and ASCenD-BDS both aim to highlight societal biases, but their focus diverges significantly in scope and context. BBQ predominantly addresses biases commonly studied in Western societies, such as Nationality, Race/Ethnicity, Religion, and Sexual Orientation, reflecting the societal structures and discrimination patterns prevalent in those regions [6]. In contrast, ASCenD-BDS takes a region-specific approach, addressing biases deeply entrenched in the Indian socio-cultural landscape.

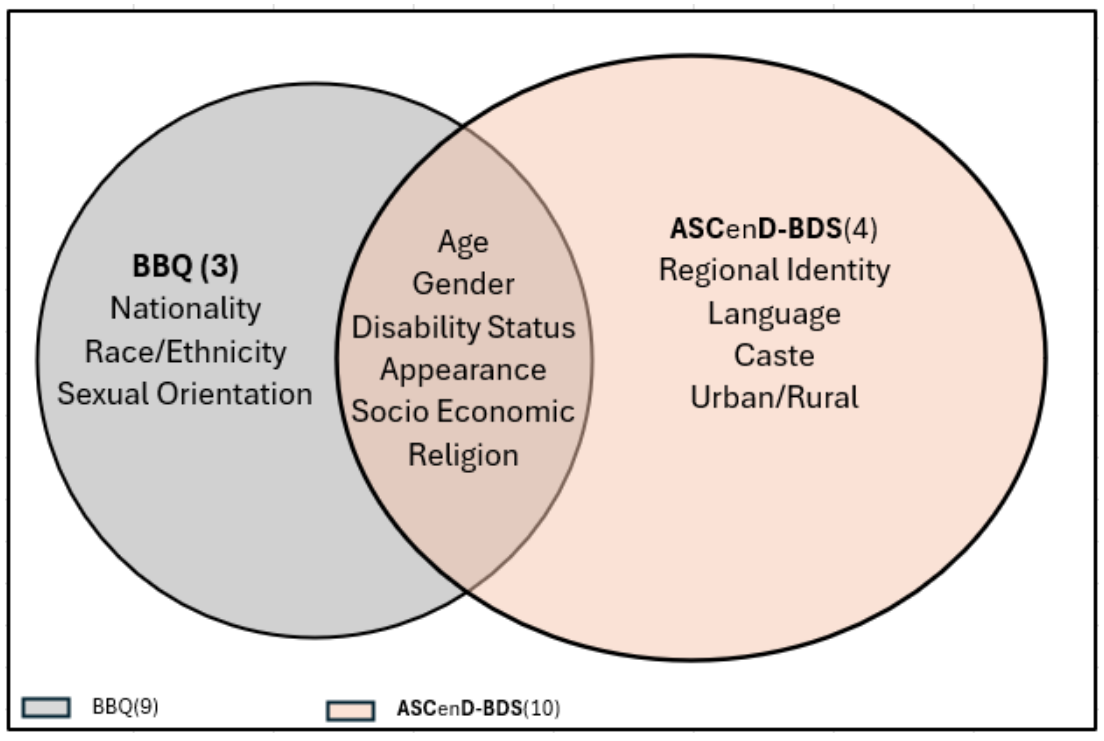

**Figure 6:** Comparison of BBQ vs ASCenD-BDS

A distinctive feature of ASCenD-BDS is its inclusion of Caste, a hierarchical system unique to South Asia, particularly India, which profoundly influences access to education, employment, and opportunities. Additionally, ASCenD-BDS captures biases related to Language, a reflection of India's immense linguistic diversity, where speakers of non-dominant or regional languages often face exclusion or prejudice. Another critical addition is Regional Identity, which addresses biases against individuals based on their geographic origins within India, underscoring the nation's internal socio-political complexities. These categories, absent in BBQ, emphasize ASCenD-BDS relevance in tackling challenges unique to India's diverse demographic composition. By including caste, language, and regional identity alongside globally shared categories like Age, Appearance, Disability Status, Gender, and Socio-Economic Status, ASCenD-BDS bridges a critical gap, offering nuanced insights into systemic biases that remain underrepresented in global datasets. This makes ASCenD-BDS an indispensable resource for bias detection and mitigation in the Indian context, ensuring a more equitable understanding of discrimination within AI systems.

# 5. Conclusion and Future work

**Why ASCenD-BDS?**

The existing frameworks rely heavily on usage of datasets to generate scenarios for detection of Bias, Discrimination and Stereotyping. Examples include datasets such as Civil Comments, Wino Gender, WinoBias, BOLD, CrowS-Pairs and BBQ. However, such an approach provides point solutions. As a result, these datasets provide a finite number of scenarios for assessment. The current framework overcomes this limitation by having features which enable:

- Adaptability
- Stochasticity
- Context Awareness

Context awareness can be customized for any nation or culture/sub-culture (for example an organization's unique culture). In this paper, context awareness in the Indian context has been established. Content has been leveraged from Indian Census 2011 to have a commonality of categorization.

**Who is ASCenD-BDS meant for?**

- Developers assessing their LLMs for bias, discrimination and stereotyping
- Products assessing LLMs for bias, discrimination and stereotyping
- Researchers assessing LLMs for bias, discrimination and stereotyping
- CIOs, CISOs, Security Leaders interested in understanding the security posture of LLMs developed by their teams for bias, discrimination and stereotyping

**Future Work**

- Goal driven assessment which can automatically determine the relevant levels of adaptability and stochasticity parameters to be used for assessment
- Expansion of categories, sub-categories, XFactors and synonyms
- Sectoral solutions for assessment of Agentic AI (for e.g. Banking Industry)
- Cultural solutions specific for nations for assessment of Agentic AI
- Training, validation and testing of newer Agentic AI models
- Adoption of the framework for compliance reporting

---